# Effective Out-of-Distribution Detection in Classifier Based on PEDCC-Loss

Qiuyu ZHU, Guohui ZHENG, Yingying YAN

School of Communication and Information Engineering, ShangHai University, 99 ShangDa Road, BaoShan District ShangHai City, China

College of Sciences, ShangHai University, 99ShangDa Road, BaoShan District ShangHai City, China.

**Abstract**

Deep neural networks suffer from the overconfidence issue in the open world, meaning that classifiers could yield confident, incorrect predictions for out-of-distribution (OOD) samples. Thus, it is an urgent and challenging task to detect these samples drawn far away from training distribution based on the security considerations of artificial intelligence. Many current methods based on neural networks mainly rely on complex processing strategies, such as temperature scaling and input preprocessing, to obtain satisfactory results. In this paper, we propose an effective algorithm for detecting out-of-distribution examples utilizing PEDCC-Loss. We mathematically analyze the nature of the confidence score output by the PEDCC (Predefined Evenly-Distribution Class Centroids) classifier, and then construct a more effective scoring function to distinguish in-distribution (ID) and out-of-distribution. In this method, there is no need to preprocess the input samples and the computational burden of the algorithm is reduced. Experiments demonstrate that our method can achieve better OOD detection performance.

## 1. Introduction

Deep neural networks (DNNs) have led to breakthroughs in many research areas, such as medical image processing, autonomous driving, and speech recognition. In classification tasks, state-of-the-art convolutional neural networks can achieve comparable or even higher accuracy than manual classification [1, 2]. However, it also exposed some problems [3]. As far as the classification task is concerned, the convolutional neural network has the problem of overconfidence, that is, misjudging OOD as ID. This will bring catastrophic losses in some cases, especially in the field of AI safety [4-6] such as unmanned driving. In this case, instead of blindly classifying, we hope that the neural network can reject OOD samples and alert users. Lacking an assessment of the uncertainty of deep neural networks will make it difficult for people to trust the predictions of DNN models [7]. This issue leads to a very crucial research direction, namely out-of-distribution detection.

Classification networks are trained under the assumption that the data fed into the neural network at test time will have the same distribution as the training data. However, this assumption does not hold in reality, as the neural network is exposed to a dynamic and variable environment, and the data input to the model will have a large offset from the training data. Actually, the data fed into the classification network can be divided into two parts, in-distribution (ID) and out-of-distribution (OOD). The ID data has the same distribution as the training data of the classification network, while the distribution of OOD data is completely different from the data used to train the neural network. Because ID data and OOD data coexist in the real world, this brings uncertainty to the neural network. Since Hendrycks et al. proposed the use of neural networks for OOD detection, DNN-based OOD detection has become one of the research hotspots, and many excellent algorithms have emerged. From different perspectives, OOD detection can also be called uncertainty estimation [8, 9], anomaly detection [10] or novelty detection [11].

In this paper, we propose an efficient method to detect out-of-distribution samples utilizing PEDCC-Loss. We explore the essence of OOD detection by decomposing the confidence score in the Euclidean space to investigate strategies improving the effectiveness of OOD detection algorithm. Our work will not involve the input preprocessing that widely used in many previous algorithms because of the computational complexity to the algorithm it will bring, and we will further discuss this theoretically in Section 2. Also, our method does not need fine-tuning for each OOD dataset, which makes our method a stable algorithm. Our main contributions are as follows:

1) A unified OOD detection framework based on PEDCC-Loss is established, which can use any pre-trained PEDCC-Loss neural network classifier to detect OOD samples without other re-training.
2) Based on mathematical analysis and derivation in Euclidean space, we decompose the confidence score output by the PEDCC-Loss classifier, and construct a more effective scoring function for OOD detection using linear weighted sum method.
3) Without complicated preprocessing methods, our proposed method has lower computational complexity. Extensive experiments illustrate our method has better OOD detection performance.

## 2. Related Work

In this section, we describe some popular algorithms that use neural networks for OOD detection. We have a dataset $D = \{(x_i, y_i)\}_{i=1}^{N}$, $x_i$ is an input data and $y_i \epsilon \{1 \dots C\}$ is its label. The goal of OOD detection using DNNs is to find a suitable scoring function S(x), where the higher score s from S(x), the higher the probability of ID. This is a binary decision, which can be easily made by applying a threshold on s. The criteria for threshold selection are based on the application requirement or the performance metric calculation protocol [15].

Hendrycks et al. proposed a baseline method for OOD detection using DNNs [12], which uses the max value of class posterior probabilities output from a Softmax-based classifier as the scoring function to classify OOD and ID. They observed that the classification network will give a higher prediction probability to the ID than OOD. The scoring function is as follows:

$$S_{base}(x) = \max_i P(y|x) \qquad (1)$$

$P(y|x)$ is the output from a Softmax-based classifier. The baseline pioneered the use of DNNs to detect OOD, but its limited performance is its biggest limitation.

Later, an improved version of the baseline, ODIN [13], was published. And two new technologies, namely temperature scaling and input preprocessing, were proposed to further enlarge the gap between OOD and ID on Softmax posterior probabilities, and ODIN achieved better results than the baseline. The input preprocessing proposed by ODIN has been widely adopted by many subsequent OOD detection algorithms, proving that it can

indeed significantly improve the performance of OOD algorithms. The input preprocessing is given by:

$$\hat{x} = x - \varepsilon sign(-\nabla_x S(x)) \quad (2)$$

The scoring function of ODIN can be described as:

$$S_{ODIN}(\hat{x}) = \max_i P(y|\hat{x}, T) \quad (3)$$

Compared with Eq. (1), $T$ is the hyperparameter for temperature scaling and $\hat{x}$ is the image after input preprocessing. Despite ODIN has made some progress, ODIN needs to tune hyperparameter $\varepsilon$ for each OOD dataset, which makes ODIN a data-sensitive algorithm. And in the real environment, it is impossible to obtain representative OOD data, which brings great instability to the performance of ODIN.

Lee K et al. [14] use the Mahalanobis distance as the OOD decision threshold, and also achieved good results. The confidence score based on Mahalanobis distance:

$$S^l_{Maha}(x) = \max_i -(f^l(x) - \mu_i^l)\sum_i^l (f^l(x) - \mu_i^l) \quad (4)$$

$$S_{Maha}(x) = \sum_l \alpha^l S^l_{Maha}(x) \quad (5)$$

The $f^l(x)$ represents the latent features at the $l$th-layer of neural networks, and $\mu_i^l$ is the class mean representation and $\Sigma$ is the covariance matrix. Mahalanobis adopts the same input preprocessing as ODIN, which also needs OOD data to tune hyperparameters, just like ODIN.

By proposing to decompose the confidence score as well as a modified input preprocessing method, Hsu et al. [15] further extended ODIN and designed a method called Generalized ODIN (G-ODIN). And G-ODIN has achieved the best results so far. The scoring function can be described as:

$$f_i(x) = \frac{h_i(x)}{g(x)} \quad (6)$$

$$S_{G-ONID}(x) = \max_i h_i(x) \text{ or } g(x) \quad (7)$$

$f_i(x)$ is the logit for class i, $h_i(x)$ and $g(x)$ are the decomposed confidence scores. Not only does G-ODIN achieve better results, but a big improvement of G-ODIN is that fine-tuning for each OOD dataset is no longer required, thus guaranteeing the stability of G-ODIN.

**About input preprocessing:** The above algorithms all require input preprocessing except for the baseline. According to Eq. (2), the input preprocessing consists of three steps, feeding the image x into a classification model to obtain S(x), then calculating the gradient at the input by back propagation, and finally feeding perturbed images $\hat{x}$ into the network again to get $S(\hat{x})$. It can be found that this process requires two forward propagations and one back propagation. For simple networks, it will not have much impact, but for complex deep networks, it is necessary to consider whether the detection speed meets the requirements. The method proposed in this paper requires only one forward propagation to obtain the scoring function, which is three times more efficient than the above algorithms.

## 3. Approach
### 3.1 PEDCC-Loss

PEDCC [16] is the abbreviation of Predefined Evenly-Distribution Class Centroids. PEDCC algorithm can generate $C$ center points that are evenly distributed on the $D$-dimensional hypersphere. $D$ is the feature dimension and $C$ is the number of classification categories. Figure 1 visualizes the neural network model based on PEDCC, which shows that PEDCC is a regression model. The neural network will extract the $D$-dimension feature vector $f$ of each input data, and calculate the cosine distance between $f$ and each center point of PEDCC as the confidence score of classification. PEDCC-Loss [17] is the loss function based on PEDCC. The Softmax-Loss classifier mainly relies on the neural network to minimize the intra-class distance, but cannot ensure the inter-class distance is large enough. PEDCC-Loss can ensure the inter-class distance sufficiently large in the regression process. The expression of PEDCC-Loss is as follows:

$$L_{PEDCC-AM} = -\frac{1}{N}\sum_i \log \frac{e^{s \cdot (cos\theta_{y_i}-m)}}{e^{s \cdot (cos\theta_{y_i}-m)} + \sum_{j=1, j \neq y_i}^{c} e^{s \cdot cos\theta_j}} \quad (8)$$

$$L_{PEDCC-MSE} = \frac{1}{N}\sum_{i=1}^{N} ||f_i - a_{y_i}||^2 \quad (9)$$

$$L = L_{PEDCC\_AM} + \sqrt[n]{L_{PEDCC-MSE}} \quad (10)$$

Eq. (8) is the AM-Softmax loss [18], where $cos\theta$ is the cosine distance between $f$ and each center point of PEDCC, as shown in Figure 2, and $s$ and $m$ are adjustable parameters. Eq. (9) is the MSE loss, calculating the distance between the feature vector $f$ and the predefined center point of $y_i$. The addition of the two is PEDCC-Loss, where n (n>=1) is a constrain factor of the PEDCC-MSE.

Since PEDCC was proposed, it has been successfully applied in clustering [19], semi supervised learning [20], incremental learning [21] and so on.

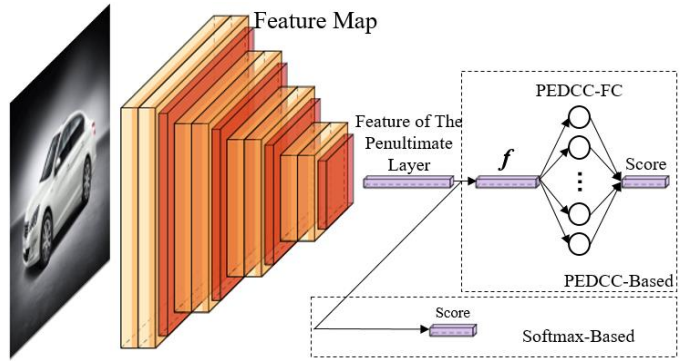

Figure 1. The difference between PEDCC-Loss and Softmax-Loss in training neural networks. PEDCC-FC is a linear layer with fixed parameters (the fixed parameters are $C$ center points), which is used to calculate the cosine distance between the feature vector $f$ and each center point.

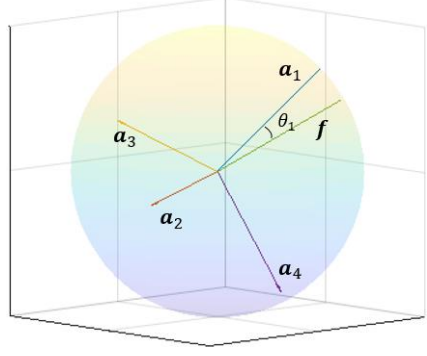

Figure 2. Visualization of the predefined evenly-distribution class centroids. For ease of visualization, we set D = 3 and C = 4. There are 4 unit vectors (**a**₁, **a**₂, **a**₃, **a**₄) evenly distributed on the 3-dimensional spherical surface. Then $y_{pred} = \underset{i}{\operatorname{argmax}} \cos\theta_i$.

### 3.2 OOD detection based on PEDCC-Loss

The baseline [12] observed that Softmax-based classifier will assign higher confidence scores to in-distribution data than out-of-distribution data. We acknowledge this view and extend this principle to our PEDCC-Loss classifier, so the PEDCC-Loss classifier will also assign higher scores to in-distribution data than out-of-distribution data. In other word, the maximum value of the confidence score output by the PEDCC classifier is similar to the max value of class posterior probabilities output by the Softmax-based classifier, and both can be applied to detect OOD samples. Then, the scoring function for distinguishing OOD and ID utilizing PEDCC-Loss can be expressed as:

$$S_{pedcc}(x) = \max_i \cos\theta_i, i = 1, \dots, C \qquad (11)$$

But like the baseline, if Eq. (11) is used directly for OOD detection, the final effect is not satisfactory.

To address the above issue, our inspiration starts from the distinctive distribution of PEDCC in Euclidean space. The PEDCC algorithm can generate the center points evenly distributed on the hypersphere to ensure the inter-class distance is large enough. This characteristic gives PEDCC many mathematical properties [22], and these properties become our breakthrough in improving OOD detection performance when we find that the confidence score output by PEDCC-based classifier can be decomposed.

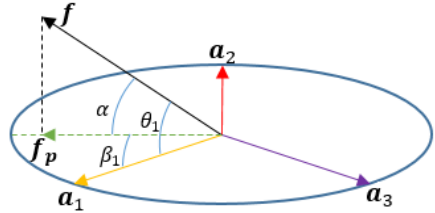

Figure 3. Taking three evenly distributed unit vectors (**a**₁, **a**₂, **a**₃) generated in a 3-dimensional space as an example.

Firstly, some properties of PEDCC discussed in our other paper [22] are as follows: For arbitrarily generated $C$ points $a_i$ $(i = 1,2,\dots,C)$ evenly distributed on the unit hypersphere of D-dimensional Euclidean space (i.e., PEDCC), and $C \leq D + 1$. α is the angle between the vector $f$ and the orthogonal projection vector $f_p$ of $f$ on the subspace spanned by the PEDCC frame. $θ_i$ and $β_i$ are the angles between $f$, $f_p$ and the *i*-th unit vector $a_i$ of the PEDCC, respectively, as shown in Figure 3. Then:

$$\cosθ_i = \cosβ_i \cdot \cosα \tag{12}$$

Eq. (12) provides us with new ideas for analyzing $S_{pedcc}$. Let $S_α(x) = \cosα$ and $S_β(x) = \max_i \cosβ_i$. Then according Eq. (12), the relationship between the two and $S_{pedcc}(x)$ is:

$$S_{pedcc}(x) = S_α(x) * S_β(x) \tag{13}$$

Eq. (13) can be interpreted as both $S_α(x)$ and $S_β(x)$ are encouraged to be larger when the data is in the high-density region of in-distribution, thus ensuring that $S_{pedcc}(x)$ reaches a larger value. So similar to $S_{pedcc}(x)$, $S_α(x)$ and $S_β(x)$ can also be used as scoring function. This means that $S_{pedcc}(x)$ is obtained by multiplying the other two scoring functions. However, because of the difference in the mean and variance of $S_α(x)$ and $S_β(x)$, Eq. (13) may amplify the effect of one and weaken the other. So we naturally think of using the linear weighted sum method to recompose the relationship of $S_α(x)$ and $S_β(x)$. Then we get a more efficient scoring function:

$$S_{D-pedcc}(x) = S_α(x) + ω * S_β(x) \tag{14}$$

$ω$ is the weight coefficient.

**Why is linear weighted sum?** We know that, if X and Y are two random variables with variances $\text{Var}(X)$ and $\text{Var}(Y)$, and $\text{Var}(X) \gg \text{Var}(Y)$, then the variance of Z=X+Y is $\text{Var}(Z) = \text{Var}(X) + \text{Var}(Y) \approx \text{Var}(X)$ (for the convenience of discussion, we assume that X and Y are independent of each other). This means that Y can be regarded as a constant with a variance of 0 relative to X, and Z=X+Y is just a translation of X. Similarly, Z=X*Y just expands X by a constant multiple. For Eq. (13), if $\text{Var}(S_β) \gg \text{Var}(S_α)$, then $S_{pedcc}$ is equal to $S_β$ multiplied by a constant, that is, the relative size of samples on $S_{pedcc}$ and $S_β$ will not change, and the OOD detection effect of $S_{pedcc}$ is almost the same as $S_β$. The linear weighted sum method can well balance the roles of $S_α$ and $S_β$ in OOD detection, realizing the effect of complementary advantages. Of course, we also tested other nonlinear combination methods, but the experimental results show that the linear combination is the best.

Designing an OOD detection algorithm learning without OOD data is a challenging task. G-ODIN achieves this for the first time by modifying the input preprocessing strategy. Through experiments, we empirically found that the $ω$ tuned with one out-of-distribution dataset could be generalized to others, so after fine-tuning $ω$ with one OOD dataset, no further tuning is needed for other OOD datasets. Therefore, as with G-ODIN, we also don't need to tune the parameters for each OOD dataset separately.

## 4. Experiments and Discussion

All our experiments use the Pytorch deep learning framework. Our code is available at: https://github.com/zhengguoh/PEDCC_OOD.

## 4.1 Experiments Settings

**In-distribution Datasets**: In our experiments, three benchmark datasets of computer vision are used, namely SVHN [23] dataset, CIFAR10 dataset and CIFAR100 dataset [24].

**Out-of-distribution Datasets**: We use the same OOD datasets as G-ODIN [15], i.e., TinyImageNet (crop), TinyImageNet (resize) [25], LSUN (crop), LSUN (resize) [26], iSUN and SVHN.

**Evaluation Metrics:** Two high-performance OOD detector evaluation metrics are used in our experiments: (1) TNR at TPR 95% can measure the true negative rate (TNR) when the true positive rate (TPR) is equal to 95%. It can be interpreted as how many OODs can be recognized as negative samples when the network recognizes 95% of ID data as positive samples. TNP at TPR 95% is the most important index to evaluate the performance of OOD detection algorithm. (2) AUROC is the Area Under the Receiver Operating Characteristic curve, which is a threshold-independent metric. The AUROC can give more reasonable results under the condition of unbalanced samples. It can be interpreted as the probability that a positive example (ID) is assigned a higher confidence score than a negative example (OOD) [27].

**Training configurations**: In all experiments, DenseNet [2] and WideResNet [28] are used as our basic classification network, and we will use PEDCC-Loss to train the classification network. The parameters in Eq. (8) and Eq. (10) are select according to the principle of the highest classification accuracy. In our experiments, we set m=5.5 and s=0.35 for cifar-10 and SVHN, and m=10 and s=0.25 for cifar-100. n in Eq. (10) is set to 1. As mentioned in 3.2, ω tuned with one out-of-distribution dataset could generalize to others. Without loss of generality, the following experiments will all use TinyImageNet(c) to tune the parameter ω.

## 4.2 Results and Discussion

**OOD benchmark performance.** In the first two experiments of Table 1, the performance of our method is compared in detail with Baseline, ODIN, Mahalanobis and G-ODIN. One can note that our method significantly outperforms the Baseline, ODIN and Mahalanobis and can be comparable to G-ODIN. In addition, our method no longer requires the use of complex input preprocessing technologies, but only requires the determination of appropriate thresholds as in the Baseline. In other words, our algorithm has the same simplicity as the Baseline, and the OOD detection effect is comparable to G-ODIN that use complex methods to obtain effective OOD detection. Besides, we added several sets of experiments to further demonstrate the effectiveness of our method, such as the last two experiments in Table 1, where other datasets and network structures are used, and we also achieved very good OOD detection results, which shows that our method is a stable and efficient algorithm. The mark "-" means that the source code and experiments are not available in the paper. We also plotted the distribution curves of OOD and ID on $S_{pedcc}(x)$ and $S_{D-pedcc}(x)$, respectively, as shown in Fig. 4 (a) and (b). One can note that the ID data on $S_{D-pedcc}(x)$ is more concentrated and the intersects area with the OOD data is less than $S_{pedcc}(x)$, which is crucial for the effectiveness of OOD detection.

Table 1: Performance of five OOD detection methods. All values are percentages and the best results are indicated in bold. Symbol ↑ means that higher scores are better.

| ID (model) | OOD | TNR at TPR 95% ↑ | AUROC ↑ |
|---|---|---|---|
| | | Baseline / ODIN / Mahalanobis / G-ODIN / $S_{pedcc}$/$S_{D-pedcc}$ | |
| CIFAR-10 (DenseNet-BC) | TinyImageNet(c) | 50.0/47.8/81.2/93.4/84.5/**97.6** | 92.1/88.2/96.3/98.7/97.8/**99.0** |
| | TinyImageNet(r) | 47.4/51.9/90.9/**95.8**/75.8/94.1 | 91.5/90.1/98.2/**99.1**/96.5/97.8 |
| | LSUN(c) | 51.8/63.5/64.2/91.5/91.9/**99.0** | 93.0/91.3/92.2/98.3/98.8/**99.6** |
| | LSUN(r) | 56.3/59.2/91.7/**97.6**/82.0/**97.6** | 93.9/92.9/98.2/**99.4**/97.5/98.9 |
| | iSUN | 52.3/57.2/90.6/97.5/83.5/**97.6** | 93.0/92.2/98.2/99.4/97.7/**99.5** |
| | SVHN | 40.5/48.7/90.6/**94.0**/81.1/93.7 | 88.1/89.6/98.0/**98.8**/97.2/97.6 |
| CIFAR-100 (DenseNet-BC) | TinyImageNet(c) | 25.3/56.0/63.5/**87.8**/42.3/**87.8** | 79.0/90.5/92.4/97.6/86.2/**98.3** |
| | TinyImageNet(r) | 22.3/59.4/82.0/**93.3**/31.1/73.8 | 76.4/91.1/96.4/**98.6**/82.7/91.0 |
| | LSUN(c) | 23.0/53.0/31.6/75.0/48.8/**93.2** | 78.6/89.9/81.2/95.3/85.6/**98.0** |
| | LSUN(r) | 23.7/64.0/82.6/**93.8**/28.7/74.6 | 78.2/93.0/96.6/**98.7**/81.4/92.4 |
| | iSUN | 21.5/58.4/81.2/**92.5**/28.2/77.4 | 76.8/91.6/96.5/**98.4**/81.4/93.5 |
| | SVHN | 18.9/35.3/43.3/77.0/31.4/**82.5** | 78.1/85.6/89.9/95.9/81.5/**97.0** |
| SVHN (DenseNet-BC) | TinyImageNet(c) | 81.3/88.5/88.6/--/92.5/**93.2** | 93.3/96.0/96.2/--/97.1/**97.6** |
| | TinyImageNet(r) | 79.8/84.1/93.0/--/96.1/**96.3** | 94.8/95.1/98.1/--/98.2/**98.3** |
| | LSUN(c) | 59.7/83.3/73.1/--/85.6/**87.2** | 89.5/93.8/94.6/--/94.8/**96.0** |
| | LSUN(r) | 77.1/81.1/91.2/--/93.5/**94.0** | 94.1/94.5/93.7/--/**97.4**/**97.4** |
| | iSUN | 73.8/93.5/92.8/--/95.8/**96.1** | 89.2/97.3/96.6/--/98.1/**98.2** |
| | CIFAR-10 | 69.3/71.7/57.7/--/88.5/**89.7** | 91.9/91.4/89.5/--/96.0/**96.6** |
| CIFAR-10 (WRN-28-10) | TinyImageNet(c) | 61.1/76.6/86.4/--/74.2/**91.4** | 92.9/94.2/96.0/--/95.4/**96.6** |
| | TinyImageNet(r) | 54.4/74.5/82.0/--/63.0/**83.6** | 91.0/92.1/94.6/--/92.5/**95.0** |
| | LSUN(c) | 65.0/78.2/78.1/--/86.5/**95.8** | 94.5/95.9/94.3/--/98.1/**98.5** |
| | LSUN(r) | 65.0/82.4/90.0/--/76.3/**90.7** | 93.9/95.4/96.5/--/95.6/**97.2** |
| | iSUN | 59.4/78.3/87.8/--/74.5/**91.0** | 92.5/93.7/96.8/--/94.4/**97.8** |
| | SVHN | 45.3/82.5/84.9/--/73.5/**90.3** | 88.9/96.7/96.1/--/95.6/**96.5** |

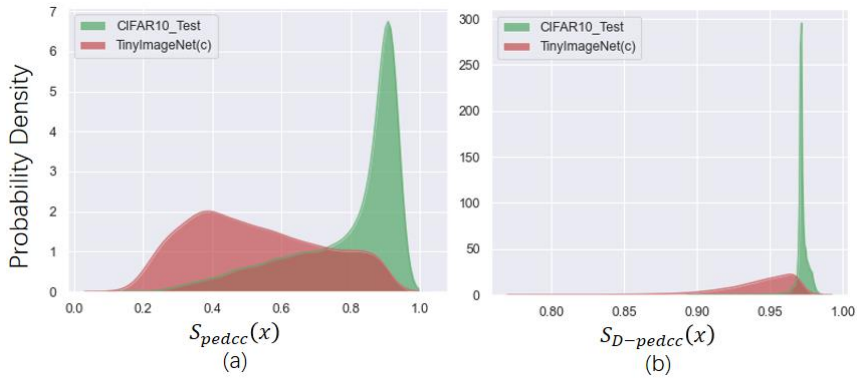

*Figure* 4. Comparison of distribution curves. (a) and (b) show the distribution of ID and OOD data on $S_{pedcc}(x)$ and $S_{D-pedcc}(x)$, respectively. The ID data used in the two images is the test data of CIFAR10. The OOD data are all TinyImageNet(c).

Table 2: A comparison about the variances between $S_\alpha$ and $S_\beta$. The data in the table are the values of variances. We calculate the variances using DenseNet-BC trained on CIFAR-10.

| Dataset | Var($S_\alpha$) | Var($S_\beta$) |
|---|---|---|
| Cifar-10 | $3.94 \times 10^{-10}$ | $2.80 \times 10^{-3}$ |
| TinyImageNet(c) | $1.83 \times 10^{-7}$ | $2.76 \times 10^{-2}$ |
| TinyImageNet(r) | $9.56 \times 10^{-8}$ | $2.44 \times 10^{-2}$ |

Table 3: Compare the OOD detection performance of $S_\beta(x)$, $S_\alpha(x) * S_\beta(x)$ and $S_\alpha(x) + S_\beta(x)$. We measure the detection performance using DenseNet-BC trained on CIFAR-10. All values are percentages.

| OOD | TNR at TPR 95% | | | |
|---|---|---|---|---|
| | $S_\alpha(x)$ | $S_\beta(x)$ | $S_\alpha(x) * S_\beta(x)$ | $S_\alpha(x) + S_\beta(x)$ |
| TinyImageNet(c) | 96.19 | 72.99 | 73.02 | 73.02 |
| TinyImageNet(r) | 91.40 | 63.11 | 63.15 | 63.14 |

**Variances vs. Detection performance.** In Section 3.2, we give a theoretical analysis for recomposing the relationship between $S_\alpha(x)$ and $S_\beta(x)$ in a linearly weighted manner. A series of experimental results are also consistent with our proposed theory. For example, the variances of $S_\alpha$ and $S_\beta$ are compared in Table 2, with $\text{Var}(S_\beta) \gg \text{Var}(S_\alpha)$. In Table 3, it can be found the OOD detection performance of $S_\beta(x)$, $S_\alpha(x) * S_\beta(x)$ and $S_\alpha(x) + S_\beta(x)$ is basically equal. This is precisely because $\text{Var}(S_\beta) \gg \text{Var}(S_\alpha)$, so that $S_\alpha$ can be regarded as a constant relative to $S_\beta$. Therefore, we choose to recompose the relationship between $S_\alpha(x)$ and $S_\beta(x)$ with a linear weighted sum method to balance the effects of the two.

Table 4: Classification accuracy on Cifar-10 and Cifar-100 datasets. The network structure is ResNet50, and all values are percentages.

| Dataset | Classification Accuracy | |
|---|---|---|
| | Softmax-Loss | PEDCC-Loss |
| Cifar-10 | 93.62 | 94.01 |
| Cifar-100 | 75.46 | 75.58 |

**Accuracy**: We propose a unified OOD detection framework based on PEDCC-Loss, capable of using any pre-trained PEDCC-Loss neural network classifier to detect OOD samples without other re-training. So our algorithm will not affect the classification accuracy of the model based on PEDCC-Loss, maintaining the advantage of PEDCC-Loss over Softmax-Loss in classification, as shown in Table 4.

**Ablation Study.** Table 5 validates the contributions of $S_\alpha(x)$ and $S_\beta(x)$. It can be found that both $S_\alpha(x)$ and $S_\beta(x)$ help to improve OOD detection performance, which demonstrates the concept mentioned in 3.2 is generally effective. Eq. (13) may amplify the effect of one and weaken the other, so that the OOD detection effect is ultimately determined by the party with the larger variance. We use the linear weighted sum method to recompose the relationship of $S_\alpha(x)$ and $S_\beta(x)$ in Eq. (14), which can realize the effect of complementary advantages.

Table 5: An ablation study about $S_\alpha(x)$ and $S_\beta(x)$. We measure the detection performance using WideResNet and DenseNet-BC trained on CIFAR-10.

| OOD | $S_\alpha(x)$ | $S_\beta(x)$ | TNR at TPR 95% | |
| --- | --- | --- | --- | --- |
| | | | WideResNet | DenseNet-BC |
| TinyImageNet(c) | √ | - | 88.7 | 96.2 |
| | - | √ | 54.8 | 73.0 |
| | √ | √ | **91.4** | **97.6** |
| TinyImageNet(r) | √ | - | 80.4 | 91.4 |
| | - | √ | 42.9 | 63.2 |
| | √ | √ | **83.6** | **94.1** |
| SVHN | √ | - | 87.6 | 92.1 |
| | - | √ | 55.6 | 71.4 |
| | √ | √ | **90.3** | **93.7** |

## 5 Conclusions

This paper concentrates on the out-of-distribution detection algorithm based on PEDCC-Loss. OOD detection is a challenging and important problem. We construct an effective scoring function for OOD detection using linear weighted sum method through mathematical analysis and derivation in Euclidean space. Furthermore, our proposed method does not change the original structure of the classification network when detecting OOD, so it can efficiently detect OOD data while ensuring that the classification network recognizes ID data accurately. By comparing with other OOD detection algorithms, it can be proved that our proposed method has better performance in detecting OOD data, and the operation is simpler. The research work in this paper can be regarded as a new baseline using the PEDCC-Loss to detect OOD. We will study and improve the algorithm based on the current results in the future, and explore the application of our method in other fields, such as active learning, incremental learning, and Stable Learning [29].

## References

[1] He, Kaiming, Zhang, Xiangyu, Ren, Shaoqing, and Sun, Jian. Deep residual learning for image recognition. In CVPR, 2016.

[2] G. Huang, Z. Liu, L. Van Der Maaten, and K. Q. Weinberger, "Densely connected convolutional networks," in Proceedings of the IEEE conference on computer vision and pattern recognition, pp. 4700–4708, 2017

[3] K. Lee, H. Lee, K. Lee, and J. Shin, "Training confidence-calibrated classifiers for detecting out-of-distribution samples," arXiv preprint arXiv:1711.09325, 2017.

[4] I. Evtimov, K. Eykholt, E. Fernandes, T. Kohno, B. Li,A. Prakash, A. Rahmati, and D. Song, "Robust physical-world attacks on machine learning models," arXiv preprintarXiv:1707.08945, vol. 2, no. 3, p. 4, 2017.

[5] D. Amodei, C. Olah, J. Steinhardt, P. Christiano, J. Schulman, and D. Mané, "Concrete problems in ai safety," arXiv preprint arXiv:1606.06565, 2016.

[6] M. Sharif, S. Bhagavatula, L. Bauer, and M. K. Reiter, "Accessorize to a crime: Real and stealthy attacks on state-of-the-art face recognition," in Proceedings of the 2016 acm sigsac conference on computer and communications security, pp. 1528–1540, 2016.

[7] Gawlikowski, Jakob, et al. "A Survey of Uncertainty in Deep Neural Networks." arXiv preprint arXiv:2107.03342 (2021).

[8] Andrey Malinin and Mark Gales. Predictive uncertainty estimation via prior networks. In S. Bengio, H. Wallach, H. Larochelle, K. Grauman, N. Cesa-Bianchi, and R. Garnett, editors, Advances in Neural Information Processing Systems 31, pages 7047–7058. Curran Associates, Inc., 2018.

[9] Andrey Malinin and Mark Gales. Reverse kl-divergence training of prior networks: Improved uncertainty and adversarial robustness. In Advances in Neural Information Processing Systems, pages 14520–14531, 2019.

[10] Andrews, Jerone, et al. "Transfer representation-learning for anomaly detection." JMLR, 2016.

[11] Masana, Marc, et al. "Metric learning for novelty and anomaly detection." *arXiv preprint arXiv:1808.05492* (2018).

[12] D. Hendrycks and K. Gimpel, "A baseline for detecting misclassified and out-of-distribution examples in neural networks," in 5th International Conference on Learning Representations, ICLR 2017, Toulon, France, April 24-26, 2017, Conference Track Proceedings, OpenReview.net, 2017.

[13] S. Liang, Y. Li, and R. Srikant, "Enhancing the reliability of out-of-distribution image detection in neural networks," arXiv preprint arXiv:1706.02690, 2017.

[14] K. Lee, K. Lee, H. Lee, and J. Shin, "A simple unified framework for detecting out-of-distribution samples and adversarial attacks," in Advances in Neural Information Processing Systems, pp. 7167–7177, 2018.

[15] Y.-C. Hsu, Y. Shen, H. Jin, and Z. Kira, "Generalized odin: Detecting out-of-distribution image without learning from out-of-distribution data," in Proceedings of the IEEE/CVF Conference on Computer Vision and Pattern Recognition, pp. 10951–10960, 2020.

[16] Q. Zhu and R. Zhang, "A classification supervised auto-encoder based on predefined evenly-distributed class centroids," arXiv preprint arXiv:1902.00220, 2019

[17] Zhu Q, Zhang P, Wang Z, et al. A New Loss Function for CNN Classifier Based on Predefined Evenly-Distributed Class Centroids[J]. IEEE Access, 2019, 8: 10888-10895.

[18] Wang F, Cheng J, Liu W, et al. Additive margin softmax for face verification[J]. IEEE Signal Processing Letters, 2018, 25(7): 926-930.

[19] Qiuyu Zhu, Zhengyong Wang. An Image Clustering Auto-Encoder Based on Predefined Evenly-Distributed Class Centroids and MMD Distance[J]. Neural Processing Letters, 11 Jan. 2020.

[20] Qiu-yu Zhu, Tian-tian Li. Semi-supervised learning method based on predefined evenly-distributed class centroids[J]. Applied Intelligence, 22 March 2020.

[21] Qiuyu Zhu, Zikuang He, Xin Ye. Incremental Classifier Learning Based on PEDCC-Loss and Cosine Distance. arXiv preprint arXiv:1906.04734.

[22] H. Hu, Y. Yan, Q. Zhu and G. Zheng, "Generation and Frame Characteristics of Predefined Evenly-Distributed Class Centroids for Pattern Classification," in IEEE Access, vol. 9, pp. 113683-113691, 2021, doi: 10.1109/ACCESS.2021.3083764.

[23] Y. Netzer, T. Wang, A. Coates, A. Bissacco, B. Wu, and A. Y. Ng, "Reading digits in natural images with unsupervised feature learning," 2011.

[24] A. Krizhevsky, G. Hinton, et al., "Learning multiple layers of features from tiny images," 2009.

[25] J. Deng, W. Dong, R. Socher, L.-J. Li, K. Li, and L. Fei-Fei, "Imagenet: A large-scale hierarchical image database," in 2009 IEEE conference on computer vision and pattern recognition, pp. 248–255, Ieee, 2009.

[26] F. Yu, A. Seff, Y. Zhang, S. Song, T. Funkhouser, and J. Xiao, "Lsun: Construction of a large-scale image dataset using deep learning with humans in the loop," arXiv preprint arXiv:1506.03365, 2015.

[27] Fawcett, Tom. "An introduction to ROC analysis." *Pattern recognition letters* 27.8 (2006): 861-874.

[28] S. Zagoruyko and N. Komodakis, "Wide residual networks," arXiv preprint arXiv:1605.07146, 2016.

[29] X. Zhang, P. Cui, R. Xu, L. Zhou, Y. He, and Z. Shen, "Deep_stable learning for out-of-distribution generalization," arXiv preprint arXiv:2104.07876, 2021.